\documentclass[a4paper,conference]{IEEEtran}
\ifCLASSINFOpdf
\else
\fi

\usepackage{algorithmic}
\usepackage{array}
\usepackage{mathtools}
\usepackage{amsmath,amssymb}

\usepackage{comment}
\usepackage{fixltx2e}
\usepackage{stfloats}
\usepackage{url}
\usepackage{xcolor}
\usepackage{multirow}
\usepackage{multicol}
\usepackage{dsfont}
\newcommand{\xmark}{\ding{55}}%
\newcommand{\cmark}{\ding{51}}%
\usepackage{pifont}
\usepackage{float}

\usepackage[ruled,vlined]{algorithm2e}

\begin{document}
\title{OTAdapt: Optimal Transport-based Approach For Unsupervised Domain Adaptation}

\author{Thanh-Dat Truong$^1$, $\quad$ Ravi Teja NVS Chappa$^1$, $\quad$ Xuan-Bac Nguyen$^1$, Ngan Le$^1$ \\ 
$\quad$ Ashley P.G. Dowling$^2$, $\quad$ Khoa Luu$^1$ \\
$^1$CVIU Lab, CSCE Dept. , 
$^2$Dept. of Entomology and Plant Pathology, University of Arkansas\\
\small\texttt{\{tt032, nchappa, xnguyen, thile, adowling, khoaluu\}@uark.edu}
% \and
% \IEEEauthorblockN{Homer Simpson}
% \IEEEauthorblockA{Twentieth Century Fox\\
% Email: homer@thesimpsons.com}
% \and
% \IEEEauthorblockN{James Kirk\\ and Montgomery Scott}
% \IEEEauthorblockA{Starfleet Academy\\
% San Francisco, California 96678--2391\\
% Telephone: (800) 555--1212\\
% Fax: (888) 555--1212}
}


%Telephone: (800) 555--1212, Fax: (888) 555--1212}


\maketitle

\begin{abstract}
Unsupervised domain adaptation is one of the challenging problems in computer vision.
This paper presents a novel approach to unsupervised domain adaptations based on the optimal transport-based distance. Our approach allows aligning target and source domains without the requirement of meaningful metrics across domains. In addition, the proposal can associate the correct mapping between source and target domains and guarantee a constraint of topology between source and target domains.
The proposed method is evaluated on different datasets in various problems, i.e. (i) digit recognition on MNIST, MNIST-M, USPS datasets, (ii) Object recognition on Amazon, Webcam, DSLR, and VisDA datasets, (iii) Insect Recognition on the IP102 dataset.
The experimental results show our proposed method consistently improves performance accuracy. 
Also, our framework can be incorporated with any other CNN frameworks within an end-to-end deep network design for recognition problems to improve their performance.
\end{abstract}


% \fi
\IEEEpeerreviewmaketitle

\section{Introduction}

Deep learning-based image recognition studies have been recently achieving very accurate performance in visual applications, e.g. image classification \cite{deep_vgg, deep_resnet, deep_densenet}, face recognition,  \cite{Luu_FG2011, Duong_ICASSP2011, Nguyen_2021_CVPR, duong2020vec2face, Quach_2021_CVPR}, image synthesis \cite{Duong_2017_ICCV, duong2016dam_cvpr, duong2019learning, duong2029cvpr_automatic, truong2021fastflow, duong2019dam_ijcv}, action recognition \cite{Truong_2022_CVPR, fi13080194}, semantic segmentation \cite{Huynh_2021_CVPR, le2018segmentation}. 
\textit{However, these methods assume the testing images from the same distribution as the training images, therefore, these deep learning-based models are likely to fail when performing in real data in the new domains.}
% In recent years, 
Hence, image recognition crossing domains play an important role to address the mentioned problem and has become an active topic in the research communities. Particularly, \textit{domain adaptation} \cite{i2i_adapt, adda_cvpr, udab_icml, DBLP:conf/aaai/ShenQZY18, Truong_2021_ICCV} has received much attention in computer vision. 
Domain adaptation refers to the problem of leveraging labeled data in a source domain to learn an accurate model in a target label-free domain.
The knowledge from the source domains will be learned and transferred to the target domains in a supervised or unsupervised manner. Specifically, domain adaptation tries to minimize the difference in the deep feature representation between source and target domains by minimizing the distance between the source and target distributions \cite{DBLP:conf/aaai/ShenQZY18, adda_cvpr, udab_icml}. These prior works have indicated the importance of the discrepancy between data distributions across domains. Hence, these works result in the principle approach to solve the domain adaptation problem is that we transform the feature distributions so as to make the target feature distributions closer to the source feature distributions and utilize the classifier learned in the source domain applying to the target domain.
In our paper, we also take this intuition into account and propose a novel framework that allows to minimize the differences between source and target feature distributions. Particularly, we approach to the domain adaptation problem based on optimal transport distances.

\begin{figure}[!t]
    \centering
    \includegraphics[width=1.0\columnwidth]{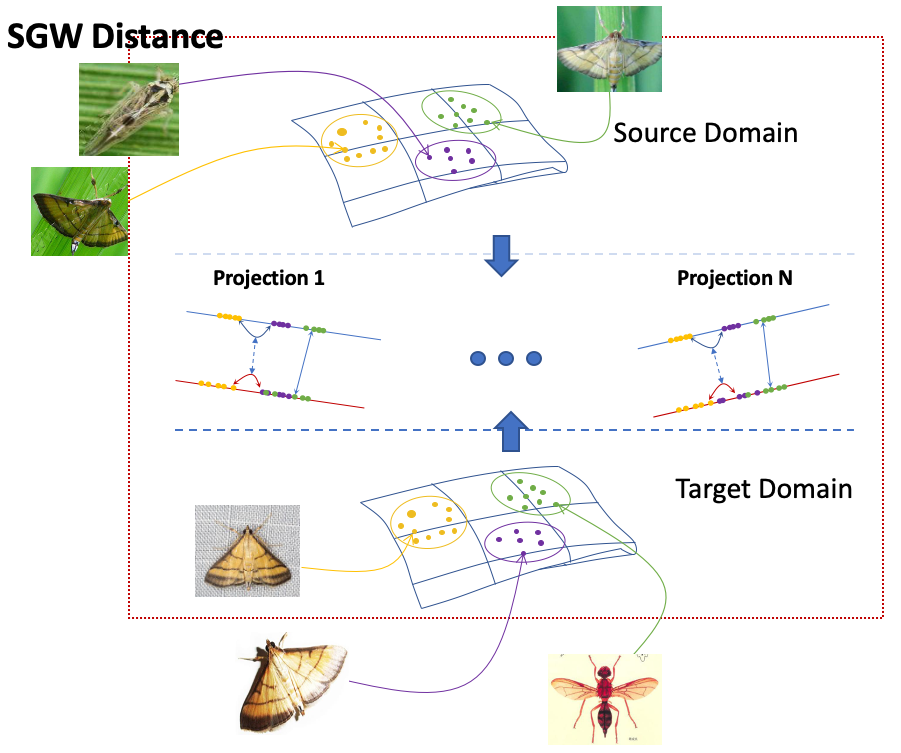}
    \caption{Optimal-Transported Based Adaptation. The Gromov-Wasserstein distance helps to align and associate target features to source features.} %\color{red}{FIX IMAGE} We propose sliced domain adaptation method based on the Sliced Gromov-Wasserstein (SGW) distance. The SGW distance helps to align and associate target features to source features.}
    \vspace{-7mm}
    \label{fig:figure_1}
\end{figure}

% Optimal Transport (OT) problems have recently raised interest in several fields, in particular because OT theory can be used for computing distances between probability distributions. Those distances, known under several names in the literature (Wasser- stein, Monge-Kantorovich or Earth Mover distances) have important properties: i) They can be evalu- ated directly on empirical estimates of the distribu-

Optimal Transport (OT) has become an active topic in recent years since it has various applications in domain adaptation \cite{pmlr-v89-redko19a, 7586038, NIPS2017_0070d23b}, generative models \cite{pmlr-v70-arjovsky17a, Deshpande_2018_CVPR, bunne2019, Truong_2021_ICCV_righ2talk}, shape matching \cite{7053911, 5457690}, etc. OT distances are used to compute the distance two probability distributions, which are known under several metrics such as $p$-Wasserstein (Earth Mover), Monge-Kantorovich, Gromov-Wasserstein distances. Theoretically, OT provides a way of inferring correspondences between two distributions by leveraging their intrinsic geometries. One of the well-known OT distances is Wasserstein which provide a way to measure two probability distributions. The Wasserstein distance is widely used for domain adaptation since it can help to mitigate the differences between source and target feature domains. However, Wasserstein leaves a serious problem, specifically, the Wasserstein distance will be not practical if we cannot define the meaningful metric across domains. In another word, if two feature domains are unaligned, we cannot directly compared or measure two data points.
To address this problem, we propose a new approach that leverages the Gromov-Wasserstein distance in deep feature spaces to compare two distributions in different domains.

% Several theoretical works [2], [36], [22] have empha- sized the role played by the divergence between the data probability distribution functions of the domains. These works have led to a principled way of solving the domain adaptation problem: transform data so as to make their distributions “closer”, and use the label information available in the source domain to learn a classifier in the transformed domain, which can be applied to the target domain. 
% Our work follows the same intuition and proposes a transformation of the source data that fits a least effort principle, i.e. an effect that is minimal with respect to a transformation cost or metric. In this sense, the adaptation problem boils down to: i) finding a transformation of the input data matching the source and target distributions and then ii) learning a new classifier from the transformed source samples. This process is depicted in Figure 1. In this paper, we advocate a solution for finding this transformation based on optimal transport.
% However, these methods often fail when the meaningful metric across domains cannot be defined.

% {\color{red}{To address this problem, we propose a new method that leverages the Sliced Gromov-Wasserstein distance in deep feature spaces to compare two distributions in different domains.}

\textbf{Contributions of this Work:}
% In order to solve this problem, we propose to use recent advanced deep learning approaches to deal with limited training samples. They include presenting novel Sliced Gromov-Wasserstein loss integrated into a deep convolutional neural network (CNN) to help to train a robust insect classification... Describe more about SGW loss, what's new? What good? How about the classification accuracy?
In order to solve the problem defined above, we propose the use of recent advanced deep learning approaches to deal with limited training samples. 
% We present a novel Sliced Gromov-Wasserstein (SGW) loss approach to domain adaptation integrated into a deep convolutional neural network (CNN) to train a robust insect classifier. 
We present a novel optimal transport loss approach with domain adaptation integrated into deep convolutional neural network (CNN) to train a robust insect classifier.
The most recent domain adaptation methods are based on adversarial training \cite{adda_cvpr, udab_icml} that minimizes the discrepancy between source and target domains. 
However, minimizing feature distributions in different domains is not practical due to the lack of a feasible metric across domains. In particular,
% In particular, 
defining a metric that is compatible with both two domains and satisfies all properties in both two domains (e.g. features of different classes should be distinguished and features of the same class should be close) is not an trivial task. 
Other prior metrics (e.g. adversarial loss, KL divergence, Wasserstein distance, etc) usually could not sufficiently satisfy this property.
Moreover, these current methods ignore the feature distribution structures between source and target domains. 
% The goal of domain adaptation is to alleviate the distribution discrepancy between two domains. However, defining the meaningful metric actress domains to reduce the difference in distribution between two domains is not an trivial task. 
% In particular, we need to define a metric that is compatible with both two domains and satisfies all properties in both two domains (e.g. features of different classes should be distinguished and features of the same class should be close). 
% Other prior metrics (e.g. adversarial loss, KL divergence, Wasserstein distance, etc) usually could not sufficiently satisfy this property.
% Meanwhile, with the proposed Gromov-Wasserstein distance, we do not need to define a metric across domains while still minimizing the distribution discrepancy across domains and maintaining the topology (relative structures) of feature distributions of two domains.
To address these mentioned issues, we propose a novel optimal transport distance, specifically, the Gromov-Wasserstein (GW) distance, that allows comparing features across domains while aligning feature distributions and maintaining the feature structures between source and target domains.
In addition, since the computation of GW distance is costly due to the solving non-convex quadratic assignment problem, we present a fast approximation form of GW distance based on 1D-GW distance.
Table \ref{tab:method_summary} summarizes the properties of our proposed method compared to other current domain adaptation methods.
Through intensive experiments on MNIST, MNIST-M, IP102, and VisDA datasets, we prove our proposed method can help to improve the performance of domain adaptation methods.

% NEED A FIGURE ON THE FIRST PAGE!

\begin{table*}[!t]
	\small
	\centering
	\caption{Comparisons in the properties between our proposed approach and other recent methods, where \xmark represents \textit{not applicable} properties. Gaussian Mixture Model (GMM), Probabilistic Graphical Model (PGM), Convolutional Neural Networks (CNN), Adversarial Loss ($\ell_{adv}$), Log Likelihood Loss ($\ell_{LL}$), Cycle Consistency Loss ($\ell_{cyc}$), Discrepancy Loss ($\ell_{dis}$) and Cross-Entropy Loss ($\ell_{CE}$), Sliced Gromov-Wasserstein Loss ($\ell_{SGW}$).}
	
% 	\resizebox{\textwidth}{!}{
	\begin{tabular}{ c|c|c|c|c|c}
% 		\Xhline{\arrayrulewidth}
		& \begin{tabular}{@{}c@{}}\textbf{Domain}\\ \textbf{Modality} \end{tabular}
		& \textbf{Network Structures}& \begin{tabular}{@{}c@{}}\textbf{Loss}\\ \textbf{Functions}\end{tabular}& 
% 		\Xhline{\arrayrulewidth}
		\begin{tabular}{@{}c@{}}\textbf{End-to-End}\end{tabular} &
% 		\begin{tabular}{@{}c@{}}\textbf{Target-domain} \\ \textbf{sample-free}\end{tabular}&
		\begin{tabular}{@{}c@{}}\textbf{Target-domain} \\ \textbf{Label-free} \end{tabular} \\
% 		&  \begin{tabular}{@{}c@{}}\textbf{Deployable} \\ \textbf{Domains}\end{tabular}\\
% 		\Xhline{\arrayrulewidth}
        % \hline
		FT \cite{feature_transfer_learning} & Transfer Learning & CNN & $\ell_{2}$
		& \cmark  & \xmark \\
		\hline
		\hline
		UBM \cite{ubm_speaker}&Adaptation & GMM & $\ell_{LL}$
		& \xmark  & \cmark \\
		DANN \cite{udab_icml}&Adaptation & CNN & $\ell_{adv}$
		& \cmark  & \cmark \\
        CoGAN \cite{cogan}&Adaptation & CNN+GAN & $\ell_{adv}$
		& \cmark &  \cmark \\
		I2IAdapt \cite{i2i_adapt} &Adaptation & CNN+GAN & $\ell_{adv}+\ell_{cyc}$
		& \cmark &  \cmark \\
		ADDA \cite{adda_cvpr} &Adaptation & CNN+GAN & $\ell_{adv}$
		& \cmark &  \cmark \\
		MCD \cite{mcd_adaptaion} &Adaptation & CNN+GAN & $\ell_{adv} + \ell_{dis}$
		& \cmark &  \cmark \\
% 		\hline
% 		\hline
		%\Xhline{1\arrayrulewidth}
% 		CrossGrad \cite{crossgrad_domain_generalization} &Generalization & Bayesian Net & $\ell_{CE}$
% 		& \cmark & \cmark & \cmark & Any \\
		ADA \cite{generalize-unseen-domain} &Generalization & CNN &$\ell_{CE}$
		& \cmark & \cmark \\
		E-UNVP \cite{e_unvp} & Generalization & PGM+CNN & ${\ell_{LL}} + {\ell_{CE}} $
		& \cmark & \cmark \\
		\textbf{OTAdapt} &\textbf{Adaptation} & \textbf{CNN + GAN} & \textbf{$\boldsymbol{\ell_{adv}} + \boldsymbol{\ell_{SGW}} $}
		& \cmark &  \cmark
% 		\textbf{Our E-UNVP} &\textbf{Generalization} & \textbf{PGM+CNN} & \textbf{$\boldsymbol{\ell_{LL}} + \boldsymbol{\ell_{CE}} $}
% 		& \cmark & \cmark & \cmark & \textbf{Any} \\
		%\textbf{} &\xmark &\xmark & \cmark & \cmark & \cmark & \cmark \\%& \cmark  \\
% 		\Xhline{\arrayrulewidth}
	\end{tabular}
% 	}
	\label{tab:method_summary}
	\vspace{-6mm}
\end{table*}

\section{Related Work}

\textbf{Domain adaptation} is a technique in machine learning, especially CNN, that aims to learn a concept from a source dataset and perform well on target datasets.
Deep convolution networks have been used in segmentation, classification, and recognition of visual domains in many applications by learning good features from the given datasets. Moreover, the learned representation from the deep convolution networks is used for other datasets. However, these representations may not generalize enough for the new datasets due to the domain shift. It is possible to mitigate this problem by fine-tuning but for large parameters employed by deep networks, it is challenging to acquire ample of labeled data.
The main goal of the domain adaptation is to reduce the discrepancy between the source and target feature distributions by leading feature learning. 

There are many works published in domain adaptation recently. The main aim of domain adaptation is to learn a distribution in a source data and find a way to improve the performance of a model on a different target data distribution. It addresses to reduce the domain shift happening between the source and the target domain. In \cite{DBLP:journals/corr/TzengHDS15}, the method maximizes domain confusion loss to learn dominant invariant representation in both source and target domains. The correlation between classes learned in the source domain transferred to target domains so that it maintains the relationship between classes.
Tzeng et. al. \cite{adda_cvpr} proposed domain adaptation using discriminative feature learning and adversarial learning for the unsupervised domain. At first, a source encoder is trained using a supervised method. Then, an adversarial adaptation is used to train the target network. Here, the discriminator that compares the source and target domain fails to recognize the difference between them. So, during testing, the trained target model with source classifier classifies the target images. 
Similarly, \cite{udab_icml} proposed a unified framework that learns the labeled data and unlabeled data at the same time.
Ber et. al. \cite{ber2020domain} presented a novel method for unsupervised domain adaptation which is suitable for imbalanced and overlapping datasets and also works with label and conditional shifts. 
% A quantification method is introduced but not well explored. A standard technique under a label shift, which employs quantification to perform domain adaptation is described. Finally, the novel scheme is derived by utilizing the quantification and domain adaptation methods. This scheme is targeted on datasets generated from electronic health records (EHRs).
Luo et. al. \cite{luo2021relaxed} identified the label-domination problem on a natural and widespread conditional GAN framework for semi-supervised domain adaptation. Also proposed Relaxed cGAN, addressing the label-domination problem by carefully designing the modules and loss functions. Here, state-of-the-art performance is obtained on Digit, DomainNet and Office-Home datasets.
Zhang et. al. \cite{zhangadversarial} proposed a novel method called Adversarial Continuous learning in unsupervised Domain Adaptation (ACDA). This proposed model confuses the domain discriminator by learning adversarially high confidence examples from the target domain. Here, a deep correlation loss is also proposed to ensure that consistency is maintained with predictions. Sener et. al. \cite{Sener:2016:LTR:3157096.3157333} proposes a unified model for learning transferable representations target label inference for unsupervised domain adaptation.

% So, to represent the learned features and train a robust classifier, they aligned both marginal and conditional distributions of source and target domains in a two-level domain alignment setting.
% \cite{udab_icml} proposed a unified framework which has a feature extractor which feeds features to both label classifier and domain classifier. During the training of source datasets, the label classifier compares the predicted value from the model and the given ground-truth value and minimize the differences to train the features. This ensures the prediction performance of the feature extractor and the label classifier on the source domain.
%and the discriminativeness of the features.
% In addition, the domain classifier used for domain invariant characteristics of the network.

% \cite{Sener:2016:LTR:3157096.3157333} proposes a unified model for learning transferable representations target label inference for unsupervised domain adaptation. In this work, the input from labeled source data and unlabelled target data feed into the framework, where the framework computes the loss alternatively between optimization of the domain transformation parameters and inferring the labels of the target domain.

\textbf{Optimal Transport} has been widely used to compute the distance between two probability distributions, which has been first introduced in middle of the 19th century.
Optimal transport has several applications in image processing (e.g. color transfer between images, etc), computer graphics (e.g. shape matching, etc). Recently, OT has gained much attention from the computer vision research society. OT has become a major metric in learning generative models \cite{pmlr-v70-arjovsky17a, Deshpande_2018_CVPR, bunne2019}, domain adaptations \cite{pmlr-v89-redko19a, 7586038, NIPS2017_0070d23b}.
However, OT suffers several issues, specifically, the computation efficiency. Computing the OT distances (e.g. Wasserstein, Gromov-Wasserstein, etc) requires a large computational cost since it has to solve the assignment problems which are NP hard problem in the general cases. Recently, there were several prior works that introduced novel methods to fast approximate the OT distances by using the sliced approaches \cite{SW_distance, NEURIPS2019_f0935e4c, vay_sgw_2019}. In our approach, we also take the intuition of the sliced approach into account to fast approximate the Gromov-Wasserstein distance.

\section{The Proposed Method} \label{sec:the_proposal}

\begin{figure*}[!t]
    \centering
    \includegraphics[width=1.5\columnwidth]{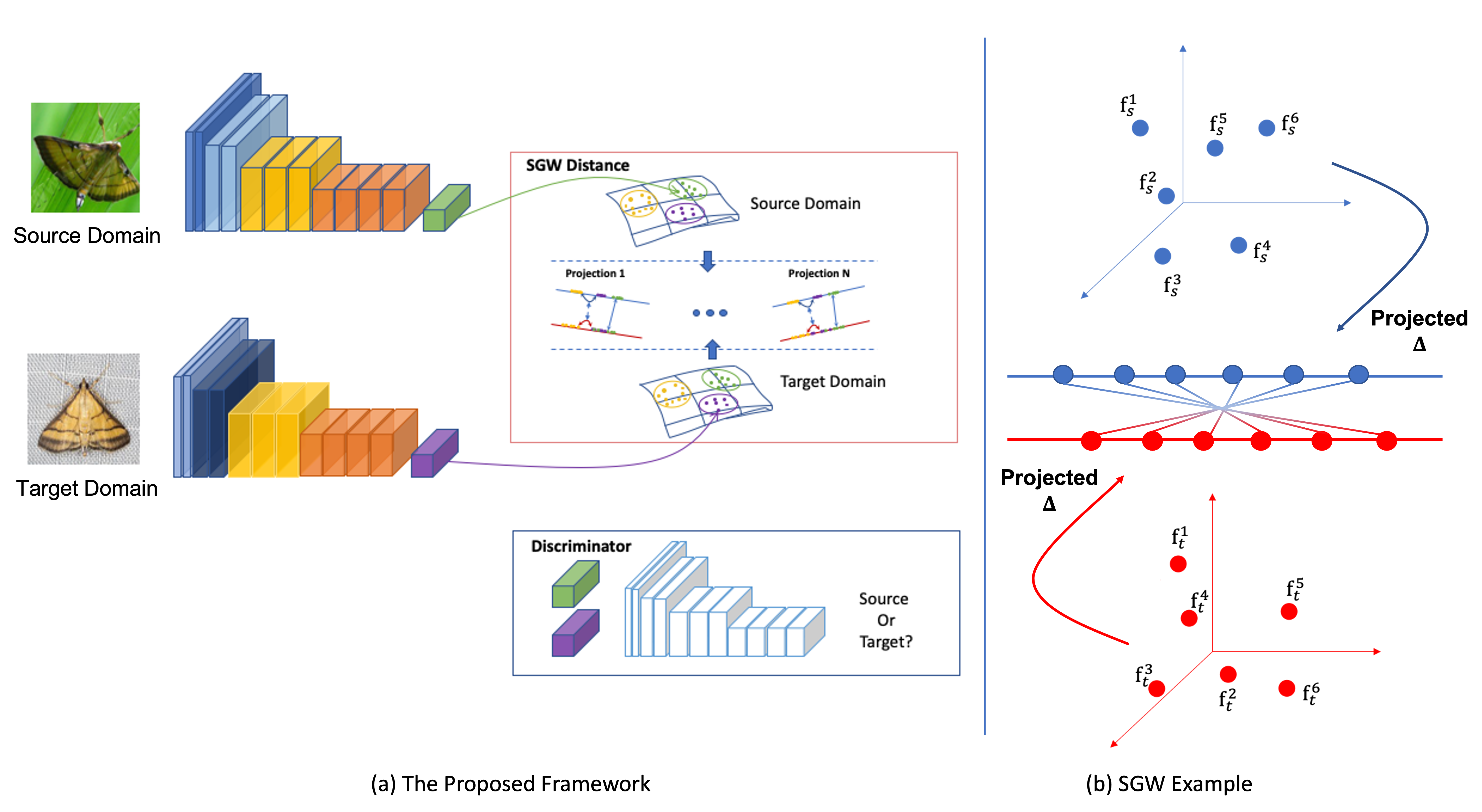}
    \vspace{-4mm}
    \caption{(a) The Proposed Framework. (b) An example of SGW in the 3D spaces that are projected to the line by a projection $\Delta$. The solution for this projection is the anti-identity mapping. }
    % \caption{The training process of our proposed method involves two main steps. First, the source model and the classifier are trained on source datasets. 
    % Then, we adapt knowledge from the pretrained source model to target model trained on the target dataset during domain adaptive training process. Finally, our method is evaluated on a new target dataset by the target model and pre-trained classifer.}
    \vspace{-6mm}
    \label{fig:training_process}
\end{figure*}

In this section, our proposed method is introduced to the problem of unsupervised domain adaptation based on the Sliced Gromov-Wasserstein distance. 
In unsupervised domain adaptation, we assume the source image $\mathbf{x}_s \in \mathbf{X}_s$ and source label $\mathbf{y}_s \in \mathbf{Y}_s$ are drawn from a source domain distribution $p_{I_s}(\mathbf{x}_s, \mathbf{y}_s)$.
Similarly, the target image $\mathbf{x}_t \in \mathbf{X}_t$ is drawn from $p_{I_t}(\mathbf{x}_t)$ and the target label $\mathbf{y_t}$ is unknown.
Fig. \ref{fig:training_process}(a) illustrates
the proposed method.
Our method aims to learn to minimize the gap between source and target distributions.
The discriminator tries to align the source and target feature representation distributions extracted from source and target extractors. Meanwhile, the Sliced Gromov-Wasserstein distance helps to associate features from the target domain to the source domain.
In other words, we try to learn a feature representation for the target domain that can utilize the classifier trained on the source domain.
% Let $\mathcal{F}_s, \mathcal{F}_t$ be the source and the target features extractors respectively, 
Let $\mathcal{F}$ be the feature extractor,
$\mathcal{C}$ be the classifier, and $\mathcal{D}$ be the discriminator.

\textbf{Network Backbone} Our network consists of two subnetworks that are a backbone network $\mathcal{F}$ and a classifier $\mathcal{C}$. 
% Symmetrically, we choose the target network are same structure with the source network. 
Particularly, we choose standard networks in our experiments, i.e. LeNet \cite{lenet_ref}, ResNet-50 \cite{deep_resnet}, VGG-16 \cite{deep_vgg} as the backbone of the source and target networks. The classifier $\mathcal{C}$ includes a fully connected layer followed by the softmax layer.
However, it should be noticed that the network structures between source and target can be different as long as the feature representations of source and target domain have the same number of dimensions.
The discriminator $\mathcal{D}$ is designed as the a stack of two fully connected layers followed by the Leaky ReLU activation.
The unsupervised domain adaptation to image classification can be formed as follows:
\begin{equation}
    \min_{\mathcal{F}, \mathcal{C}} \left[\mathop{\mathbb{E}}_{\mathbf{x}_s, \mathbf{y}_s}\mathcal{L}_s(\mathbf{x}_s, \mathbf{y}_s; \mathcal{F}, \mathcal{C}) + \mathop{\mathbb{E}}_{\mathbf{x}_t}\mathcal{L}_t(\mathbf{x}_t; \mathcal{F})\right]
\end{equation}
where $\mathcal{L}_s$ is the supervised loss on the source domain that can be defined as follows:
% \textbf{Supervised Source Domain Training:} Since we can access annotations of the source domain, the source feature extractor $\mathcal{F}_s$ and the classifier $\mathcal{C}$ can be trained using the standard supervised loss as follows,
%
\begin{equation}
    % \min_{\mathcal{F}_s, \mathcal{C}}
    \mathcal{L}_s(\mathbf{x}_s, \mathbf{y}_s; \mathcal{F}, \mathcal{C}) =
    % \mathop{\mathbb{E}}_{\mathbf{x}_s, \mathbf{y}_s}
    -\sum_{i=1}^{c}\mathds{1}_{k=\mathbb{y}_s}\log\mathcal{C}(\mathcal{F}(\mathbf{x}_s))
\end{equation}
and the $\mathcal{L}_t$ is the unsupervised loss defined on the target domain. 
% \textbf{Domain Adaptive Training:} 
Let $\mathbf{f}_s \sim p_s(\mathbf{f}_s), \mathbf{f}_t \sim p_t(\mathbf{f}_t)$ be the features extracted from the source image $\mathbf{x}_s$ and the target image $\mathbf{x}_t$ by the feature extractor $\mathcal{F}$, respectively. To adapt the knowledge from the source domain to the target domain, we minimize the source and the target feature representation by minimizing the gap between $p_s$ and $p_t$. This can be addressed by the adversarial training. The domain discriminator $\mathcal{D}$ will classify whether a feature $\mathbf{f}$ comes from the source or the target domain. $\mathcal{D}$ can be optimized by adversarial loss as follows,
\begin{equation}
\begin{split}
    % \min_{\mathcal{D}}&\mathcal{L}(\mathbf{f}_s, \mathbf{f}_t) = \mathcal{L}(\mathcal{F}_s(\mathbf{x}_s), \mathcal{F}_t(\mathbf{x}_t)) \\
    \min_{\mathcal{D}}
    % \mathcal{L}_{\mathcal{D}}(\mathbf{x}_t; \mathcal{D}) = 
    % \\
    % &= -\mathop{\mathbb{E}}_{\mathbf{x}_s \sim \mathbf{X}_s}
    \mathop{\mathbb{E}}_{\mathbf{x}_s, \mathbf{x}_t}\left[-\log \mathcal{D}(\mathcal{F}(\mathbf{x}_s))
    -
    % \mathop{\mathbb{E}}_{\mathbf{x}_t \sim \mathbf{X}_t}
    \log [1 - \mathcal{D}(\mathcal{F}(\mathbf{x}_t))]\right]
\end{split}
\end{equation}
Next, we adapt knowledge from the source feature extractor $\mathcal{F}_s$ to the target feature extractor $\mathcal{F}_t$. Hence, the target feature extractor $\mathcal{F}_t$ is optimized according to the adversarial loss as follows,
\begin{equation} \begin{split} \label{eqn:adv_loss_target}
    % \min_{\mathcal{F}_t}
    \mathcal{L}_{adv}(\mathbf{x}_t; \mathcal{F}) 
    % &= \mathcal{L}(\mathcal{F}_s(\mathbf{x}_s), \mathcal{F}_t(\mathbf{x}_t))
    % \\
    % &=-\mathop{\mathbb{E}}_{\mathbf{x}_t \sim \mathbf{X}_t}
    = -\log\mathcal{D}(\mathcal{F}(\mathbf{x}_t))
\end{split} \end{equation}

Domain adaptive training helps to minimize the distance between source and target distributions via domain adversarial training; however, it is insufficient due to three reasons: (1) adversarial training helps to align two distributions without guaranteeing the correct mapping of each classes, (2) this approach fails when a meaningful metric across domains cannot be defined, and (3) the adversarial loss ignores the topology of features distributions between two domains. To address these aforementioned issues, we adopt the optimal transport distance, i.e. the Gromov-Wasserstein distance, to mitigate the issues caused by misaligned domains.

% Gromov-Wassersein (GW) distance that can sufficiently align two distributions across domains.

% \begin{figure}[!t]
%     \centering
%     \includegraphics[width=1.0\columnwidth]{figures/GW_1D.png}
%     % \caption{Our proposed framework includes three networks: source and target feature extractors and the discriminator. The discriminator aligns the distribution of source and target features. SGW tends to associate features from target to source domains.}
%     \caption{An example of SGW in the 3D spaces that are projected to the line by a projection $\Delta$. The
%     solution for this projection is the anti-identity mapping.}
%     \label{fig:ill_of_SGW}
% \end{figure}

\textbf{Gromov-Wasserstein Distance} Let $\pi$ be a correspondence map such that $p_s$ and $p_t$ are marginal distributions of $\pi$. The distance between two distributions $p_s$ and $p_t$ across domains can be formulated as follows,
\begin{equation} \label{eqn:GW_distance}
    GW_2^2(c_{p_s}, c_{p_t}, p_s, p_t) = \min_{\pi \in \Pi(p_s, p_t)}J(c_{p_s}, c_{p_t}, \pi)
\end{equation}
where
\begin{equation}\label{eqn:J_distance}
    J(c_{p_s}, c_{p_t}, \pi) = \sum_{i,j,k,l}|c_{p_s}(\mathbf{f}_s^i, \mathbf{f}_s^j) - c_{p_t}(\mathbf{f}_t^k, \mathbf{f}_t^l)|^2\pi_{i,j}\pi_{k,l}
\end{equation}
$c_{p_s}, c_{p_t}$ are the distances in their space, in our method, we utilize squared euclidean distances, i.e. $c_{p_s}(\mathbf{f}_s^i, \mathbf{f}_s^j) = ||\mathbf{f}^i_s - \mathbf{f}_s^j||_2^2$, $c_{p_t}(\mathbf{f}_t^k, \mathbf{f}_t^l) = ||\mathbf{f}^k_t - \mathbf{f}_t^l||_2^2$.
The GW distance aims to map pairs of features with similar distances within each pair, 
specifically, 
the pairs $c_{p_s}(\mathbf{f}_s^i, \mathbf{f}_s^j)$ is associated to $c_{p_t}(\mathbf{f}_t^k, \mathbf{f}_t^l)$ when the distances are similar and the transport coefficients $\pi_{i,j}$ and $\pi_{k,l}$ of these pairs are high respond.

As shown in Eq. \eqref{eqn:GW_distance}, we only need to know the intra-distance of each domain without defining any metric across two domains. In particular, the Ecludian distance has been used as intra-distance of each domain since the Ecludian distance is invariant to permutations, rotations, or translation. This invariant-property allows GW to align the complex feature domains. In addition,  the corresponding map (transportation map) $\pi$ illustrates the association between source features and targets features, which helps to guarantee the correct mapping of each classes between two domains.
Also, the term $|c_{p_s}(\mathbf{f}_s^i, \mathbf{f}_s^j) - c_{p_t}(\mathbf{f}_t^k, \mathbf{f}_t^l)|^2$ of the Eq. \eqref{eqn:J_distance} implies the constraint of the topology of feature distributions between two domains have to be identical. Fig. \ref{fig:mnist_mnistm_distribution}(B) illustrates the aligned features distributions of source and target domains when using the Gromov-Wasserstein distance.
% the transportation map $\pi$ helps to guarantee the problem 

However, solving the equation Eq. \eqref{eqn:GW_distance} is costly due to optimizing a non-convex Quadratic Problem with the time complexity is $O(n^3)$. 
Instead of directly solving the GW distance, we present the Sliced Gromov-Wassertein (SGW) distance \cite{vay_sgw_2019} with the time complexity is less costly than GW distance.
It is similar to the Sliced Wasserstein distance \cite{SW_distance}, features are projected from the high dimensional space to the 1D space, and then solving GW distance on the 1D space. As the results of Quadratic Assignment Problem \cite{vay_sgw_2019}, solving GW on the 1D space is effectively sufficient.
Therefore, the Eq. \eqref{eqn:GW_distance} on the 1D space can be formulated as follows,
\begin{equation} \small
\begin{split} \label{eqn:GW_1D}
     GW_2^2(c_{p_s}, c_{p_t}, p_s, p_t,\Delta)=\min_{\sigma}\frac{1}{n^2}\sum_{i,j}|c_{p_s}&(\mathbf{\bar{f}}_s^i, \mathbf{\bar{f}}_s^j) \\
     &- c_{p_t}(\mathbf{\bar{f}}_t^{\sigma(i)}, \mathbf{\bar{f}}_t^{\sigma(j)})|^2
\end{split}
\end{equation}
where $\sigma$ is a one-to-one mapping $\{1,...,n\} \rightarrow \{1,...,n\}$, $\mathbf{\bar{f}}$ is a projected feature of $\mathbf{f}$ on 1D space, $\Delta$ is a projection matrix. 
% Fortunately, $\sigma$ is just either the identity mapping $\sigma(i) = i$ or anti-identity mapping $\sigma(i) = n - i$ and can be computed in $O(n\log(n))$ where $n$ is the number of data points.
Fortunately, if the source and target projected features are sorted in the increasing order, the solution for  $\sigma$ is just either the identity mapping $\sigma(i) = i$ or anti-identity mapping $\sigma(i) = n - i$. Therefore, the Eq. \eqref{eqn:GW_1D} can be computed in $O(n\log(n))$  where n is the number of data points.
Fig. \ref{fig:training_process}(b) illustrates an example of solving GW in the 1D space.

% \begin{algorithm}[!t]
% \SetAlgoLined
% \SetKwInOut{Input}{Input}\SetKwInOut{Output}{Output}
% \Input{Source Features: $\mathbf{f}_s^1, \mathbf{f}_s^2, ..., \mathbf{f}_s^n$}
% \Input{Target Features: $\mathbf{f}_t^1, \mathbf{f}_t^2, ..., \mathbf{f}_t^n$}
% \Output{The SGW distance}
%  $SGW = 0.$\;
%  $L = 200$\;
% %  \If{$d_s < d_t$}{
% %     Pad zeros to $\mathbf{f}_s$
% %  }
% %  \If{$d_s > d_t$} {
% %     Pad zeros to $\mathbf{f}_t$
% %  }
%  \For{$i\leftarrow 1$ \KwTo $L$}{
%     Random $\Delta$\;
%     $\bar{\mathbf{f}}_s = \operatorname{Project}(\mathbf{f}_s, \Delta)$\;
%     $\bar{\mathbf{f}}_t = \operatorname{Project}(\mathbf{f}_t, \Delta)$\;
%     $GW_{1D} = GW_2^2(c_{p_s}, c_{p_t}, p_s, p_t,\Delta)$\;
%     $SGW = SGW + GW_{1D}$
%  }
%  \KwRet{$\frac{SGW}{L}$}
%  \caption{Sliced Gromov-Wasserstein Distance}\label{alg:sgw_distance}
% \end{algorithm}

\textbf{Sliced Gromov-Wasserstein Distance} As aforementioned, similar to the Sliced Wasserstein (SW) distance, the main idea of SW is to project features in the high dimensional space to the 1D space where computing Wasserstein distance is simple and easy followed by averaging these distances. In SGW, the same manner is applied, specifically, the SGW distance can be defined as follows,
\begin{equation}\label{eqn:sgw_loss}
\begin{split}
    \mathcal{L}_{SGW}(\mathbf{x}_s, \mathbf{x}_t) &= SGW(c_{p_s}, c_{p_t}, p_s, p_t)\\ &=\int_{\Delta \in \mathbb{R}^{d-1}}GW_2^2(c_{p_s}, c_{p_t}, p_s, p_t,\Delta) d\Delta \\ &= \frac{1}{L}\sum_{i=1}^LGW_2^2(c_{p_s}, c_{p_t}, p_s, p_t, \Delta_i)
\end{split}
\end{equation}
% where $d = \max(d_s, d_t)$, $d_s$ and $d_t$ are 
where $d$ is the number of dimensions of source and target feature spaces; $L$ is the number of projections. 
% In case the dimensions of the source and target spaces differ, we will pad the features of the lower dimension with zeros to guarantee the dimensions between source and target spaces are equal before projecting by $\Delta$. 
In our experiments, we set the number of projections $L$ to $200$.
% The Algorithm \ref{alg:sgw_distance} illustrates steps of computing SGW distance. 
% Consequently, 
The time complexity of computing SGW distance is $O(Ln\log(n))$.

% As explained in the introduction, this result was used to approximate the Wasserstein
% distance between measures of R
% p using the so called Sliced Wasserstein (SW) distance [14]. The
% main idea is to project the points of the measures on lines of R
% p where computing a Wasserstein
% distance is easy since it only involves a simple sort and to average these distances. It has been proven
% that SW and W are equivalent in terms of metric on compact domains [13]. In the same philosophy
% we build upon Theorem 3.2 to define a ”sliced” version of the GW distance.

% , $L$ is the number of projections. 
% In our experiments, we use $L = 200$ projections.
% Consequently, the SGW loss is formulated as follows,
% \begin{equation} \label{eqn:sgw_loss}
%     \mathcal{L}_{SGW} = \frac{1}{L}\sum_{i=1}^LGW_2^2(c_{p_s}, c_{p_t}, p_s, p_t, \Delta_i)
% \end{equation}

% \textcolor{red}{DESCRIBE MORE DETAIL SGW HERE, HOW TO PROJECT N-DIMS TO 1-DIM}

Finally, the total loss for the target feature extractor is a summation of adversarial loss (Eq. \eqref{eqn:adv_loss_target}) and SGW loss (Eq. \eqref{eqn:sgw_loss}).
\begin{equation} \label{eqn:total_loss_g}
    % \min_{\mathcal{F}_t}
    \mathcal{L}_t(\mathbf{x}_t; \mathcal{F}) = \lambda_{adv}\mathcal{L}_{adv}(\mathbf{x}_t; \mathcal{F}) + \lambda_{SGW}\mathcal{L}_{SGW}(\mathbf{x}_s, \mathbf{x}_t)
\end{equation}
where $\lambda_{adv}$ and $\lambda_{SGW}$ are control weights for $\mathcal{L}_{adv}$ and $\mathcal{L}_{SGW}$, respectively. %In our experiments, we set $\lambda_{adv} = \lambda_{SGW} = 1.0$.
% 
% Fig. \ref{fig:training_process}(a) illustrates the training process of our method. 
% It includes two main steps.
% Firstly, we train a source model $\mathcal{F}_s$ and a classifier $\mathcal{C}$ on the source dataset. Then, we transfer knowledge from the source pre-trained model $\mathcal{F}_s$ to the target model $\mathcal{F}_t$ to adapt to the new target dataset by using SGW loss.
% Finally, we test our method on new target domain by the target model $\mathcal{F}_t$ and classifier $\mathcal{C}$.

\section{Experiments} \label{sec:experiments}

\begin{figure}[!t]
    \centering
    \includegraphics[width=0.9\columnwidth]{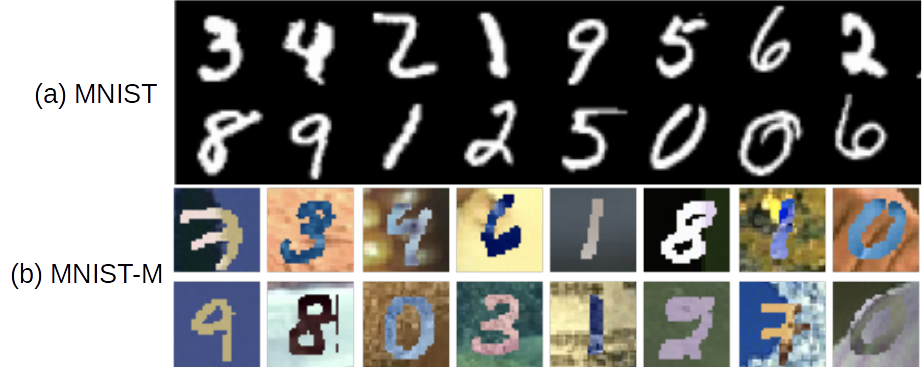}
    \vspace{-2mm}
    \caption{Examples of MNIST and MNIST-M datasets}
    % \vspace{-6mm}
    \label{fig:mnist_dataset}
\end{figure}

\begin{table}[!t]
    \small
    \centering
    % \caption{Effectiveness of Loss}
    \vspace{-4mm}
    \caption{Ablative experiment results (\%) on the effectiveness of the adversarial loss ($\mathcal{L}_{adv}$) and Gromov-Wasserstein loss ($\mathcal{L}_{SGW}$). We evaluate our proposed method in the cases of MNIST $\to$ MNIST-M and MNIST-M $\to$ MNIST.}
    \begin{tabular}{|c|c|c|}
        \hline
        Methods &  MNIST $\to$ MNIST-M &  MNIST-M $\to$ MNIST\\
        \hline
        % Pure-CNN & 62.42\% &98.45\%  \\
        Pure-CNN & 58.49\% &98.45\%  \\
        \hline
        $\mathcal{L}_{adv}$ Only & 64.77\% & 63.26\% \\
        \hline
        $\mathcal{L}_{SGW}$ Only & 65.72\% & 99.06\%\\
        \hline
        $\mathcal{L}_{adv} + \mathcal{L}_{SGW}$ & \textbf{68.56\%}  & \textbf{99.19\%} \\
        \hline
    \end{tabular}
    \vspace{-4mm}
    \label{tab:effect_loss}
\end{table}

In this section, we first show the impact of our proposed method compared to other methods in Sec \ref{sec:ablation_study}. In these experiments, we consider MNIST as the source dataset and MNIST-M as the target dataset. 
The proposed method is also benchmarked on different network structures, 
% i.e. LeNet \cite{lenet_ref}, MobileNet V2 \cite{mobilenet_v2}, VVG \cite{deep_vgg}, ResNet \cite{deep_resnet}, DenseNet \cite{deep_densenet}.
i.e. LeNet \cite{lenet_ref}, VGG \cite{deep_vgg}, ResNet \cite{deep_resnet}.
Finally, we show the advantages of our method in the across-domain pest insect recognition on IP102 dataset \cite{IP_102_dataset} in Sec \ref{label:insect_results}. In our experiments, the accuracy metric is used to compare our method and prior approaches.

\subsection{Ablation Studies} \label{sec:ablation_study}

This ablation study aims to compare our method against to other domain adaptation methods. In these experiments, MNIST and MNIST-M are used as the source and the target datasets, respectively. Fig. \ref{fig:mnist_dataset} illustrates samples of MNIST and MNIST-M datasets.
We compare our proposed method (SGW) against to Pure-CNN, ADDA \cite{adda_cvpr}, ADA \cite{generalize-unseen-domain}, TCA \cite{TCA_method}, SA \cite{SA_method}, DAN \cite{DAN_method}, UNVP, and E-UNVP \cite{e_unvp}.

\begin{table}[!b]
    \centering
    \vspace{-7mm}
    \caption{Experimental results on MNIST $\to$ MNIST-M.}
    % \vspace{-4mm}
    \label{tab:mnist2mnistm}
    \begin{tabular}{|c|c|c|}
        \hline
        Method          & MNIST   & MNIST-M \\
        \hline
        Pure CNN        & 99.33\% & 58.49\% \\
        \hline
        SA \cite{SA_method} & 90.80\% & 59.90\%\\
        \hline
        DAN \cite{DAN_method}     & 97.10\% & 67.00\% \\
        \hline
        TCA \cite{TCA_method}     & 78.40\% & 45.20\%\\
        \hline
        ADA \cite{generalize-unseen-domain} & 99.17\% & 60.02\% \\
        \hline
        ADDA \cite{adda_cvpr} & 99.29\% & 63.39\% \\
        \hline
        UNVP \cite{e_unvp} & 99.30\% & 59.45\% \\
        \hline
        E-UNVP \cite{e_unvp} & \textbf{99.42\%} & 61.70\% \\
        \hline
        \textbf{OTAdapt}    & 99.19\% & \textbf{68.56\%} \\
        \hline
    \end{tabular}
    % \vspace{-2mm}
\end{table}

\textbf{Hyper-parameter Settings:} During the training, the batch size and the learning rate are set to $128$ and $0.0002$, respectively. For the control weights $\lambda_{adv}$ and $\lambda_{SGW}$ in Eqn \eqref{eqn:total_loss_g}, we set $\lambda_{adv} = \lambda_{SGW} = 1.0$.For the training processes, we train $10$ epochs for each process. We use image sizes $32\times32$ for LeNet and $64\times64$ for VGG and ResNet.

As shown in Table \ref{tab:effect_loss}, the proposed $\mathcal{L}_{SGW}$ and $\mathcal{L}_{adv}$ help to improve the accuracy of the network on target dataset. When both $\mathcal{L}_{adv}$ and $\mathcal{L}_{SGW}$ are adopted, the performance of the proposed method is significantly improved.
Table \ref{tab:mnist2mnistm} illustrates our results compared to other methods. In this experiment, LeNet is used for all methods in the table.
As shown in the results, our method can achieve the state-of-the-art performance and help to improve performance of the model from  %\textcolor{red}
{$58.49\%$ to $68.56\%$} on MNIST-M datasets. The experimental results have shown that with our approach, the performance of the model has been improved on the color images (MNIST-M). However, although the model has been generalized into a new color image domain, there is a minor decrease in the performance of the gray-scale images (MNIST).

\textbf{Deep Network Structures} This experiment evaluates the robustness and consistent improvement of our method with common deep networks, including, LeNet, VGG, ResNet. The proposed method consistently outperform than the stand-alone deep network (Pure-CNN). As shown in Table \ref{tab:networks}, the proposed method helps to improves %\textcolor{red}
{$10.07\%$, $4.31\%$, $3.21\%$} on MNIST-M using LeNet, VGG, ResNet, 
respectively.

\begin{figure}[!t]
    \centering
    \vspace{-4mm}
    \includegraphics[width=0.45\textwidth]{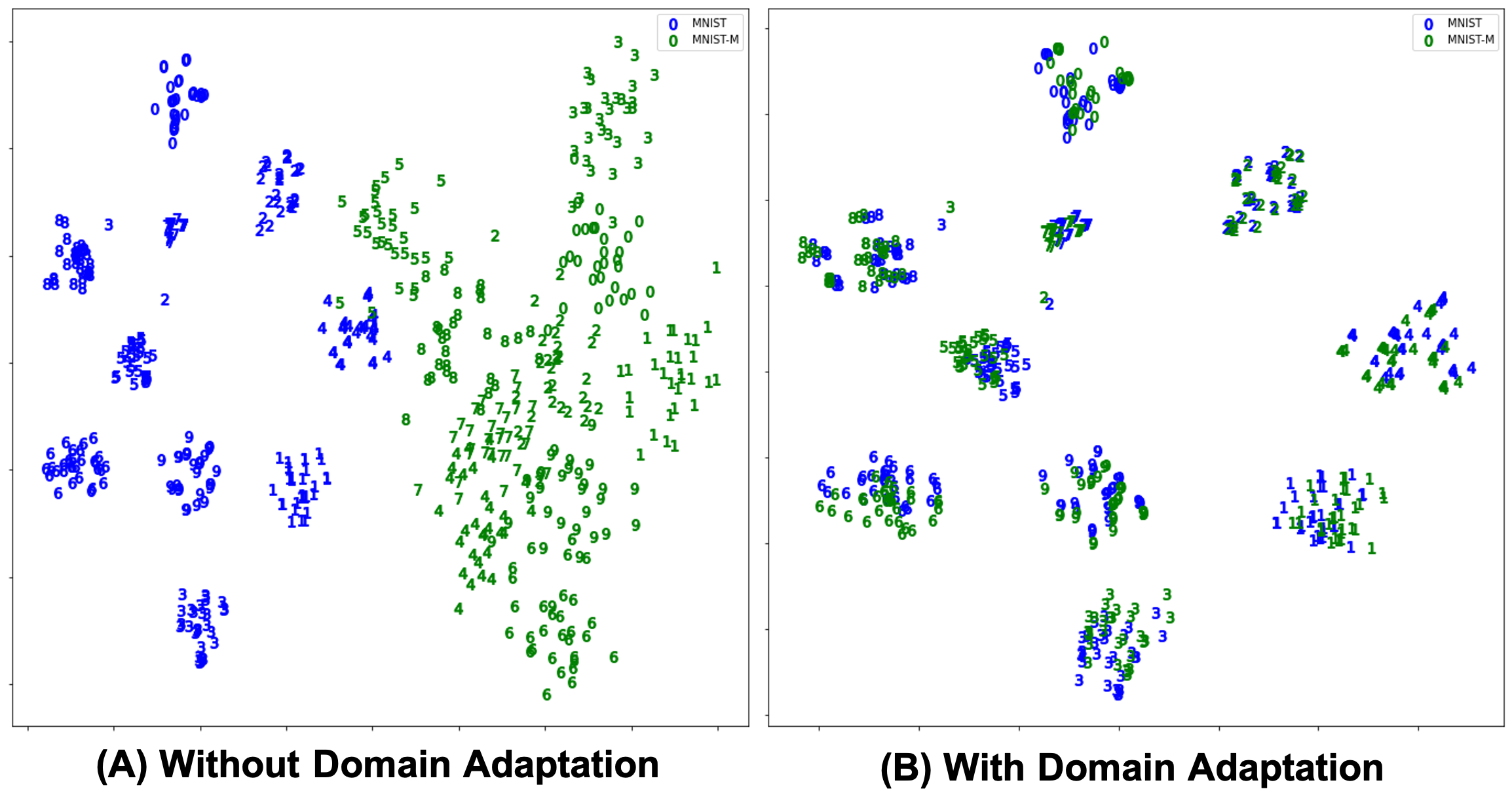}
    \vspace{-4mm}
    \caption{Feature Distributions of MNIST and MNIST-M.}
    \vspace{-6mm}
    \label{fig:mnist_mnistm_distribution}
\end{figure}

\textbf{Sample Distributions}
Fig. \ref{fig:mnist_mnistm_distribution} illustrate the feature distributions of MNIST (source dataset) and MNIST-M (target dataset) in the cases of with domain adaptation and without domain adaptation. Features of 10 classes extracted from testing sets of MNIST (blue points) and MNIST-M (green points) are projected into the 2D space by the t-SNE method. 
As shown in Fig. \ref{fig:mnist_mnistm_distribution}(A), the features of MNIST-M are not well distributed. Meanwhile, features of MNIST and MNIST-M visualized on Fig. \ref{fig:mnist_mnistm_distribution}(B) are well aligned.

\begin{table}[!t]
    \small
    \centering
    % \vspace{-6mm}
    \caption{Experimental results  $(\%)$ when using SGW in various common CNNs on MNIST $\to$ MNIST-M.} % {\textcolor{red}{@RAVI, Please update this table}}}
    \label{tab:networks}
    
    % \begin{tabular}{|c|c|c|c|}
    %     \hline
    %     \textbf{Networks} & \textbf{Methods} & \textbf{MNIST} & \textbf{MNIST-M} \\
    %     \hline
    %     \multirow{2}{*}{LeNet}      & Pure-CNN          & 99.33\%  &  58.49\% \\
    %     \cline{2-4}
    %     \multirow{2}{*}{}           &  $\ell_{adv} + \ell_{SGW}$   & \textbf{99.33\%} & \textbf{67.64\%} \\
    %     \hline
        
    %     \multirow{2}{*}{VGG}        & Pure CNN & 99.15\% & 44.08\% \\
    %     \cline{2-4}
    %     \multirow{2}{*}{}           &  $\ell_{adv} + \ell_{SGW}$   & \textbf{99.15\%}  & \textbf{75.16\%} \\
    %     \hline
        
    %     \multirow{2}{*}{ResNet}     & Pure CNN          &  99.11\% & 40.77\% \\
    %     \cline{2-4}
    %     \multirow{2}{*}{}           &  $\ell_{adv} + \ell_{SGW}$   &  \textbf{99.11\%} & \textbf{60.94\%} \\
    %     \hline
    % \end{tabular}
    
    \begin{tabular}{|c|c|c|c|}
        \hline
        \textbf{Networks} & \textbf{Methods} & \textbf{MNIST} & \textbf{MNIST-M} \\
        \hline
        \multirow{2}{*}{LeNet}      & Pure-CNN      & \textbf{99.33\%} & 58.49\% \\                
        % & 98.57\%&59.24\% \\
        \cline{2-4}
        \multirow{2}{*}{}           &  \textbf{OTAdapt}            & 
         99.19\% & \textbf{68.56\%} \\
        % \textbf{98.80\%}&\textbf{60.07\% }\\
        \hline
        
        \multirow{2}{*}{VGG}        & Pure CNN                      &98.91\%&60.95\%
        \\
        \cline{2-4}
        \multirow{2}{*}{}           &  \textbf{OTAdapt}            & \textbf{99.00\%} & \textbf{65.26\%} \\
        \hline
        
        \multirow{2}{*}{ResNet}     & Pure CNN                     &98.97\% & 64.23\% \\
        \cline{2-4}
        \multirow{2}{*}{}           &  \textbf{OTAdapt}            &\textbf{99.31\%} & \textbf{67.44\%} \\
        \hline
    \end{tabular}
\end{table}

% \vspace{-6mm}

\subsection{Insect Pest Recognition} \label{label:insect_results}

% \textcolor{red}{MORE DETAIL IP102 DATASET HERE}
\begin{figure}[!t]
    \centering
    \includegraphics[width=0.7\columnwidth]{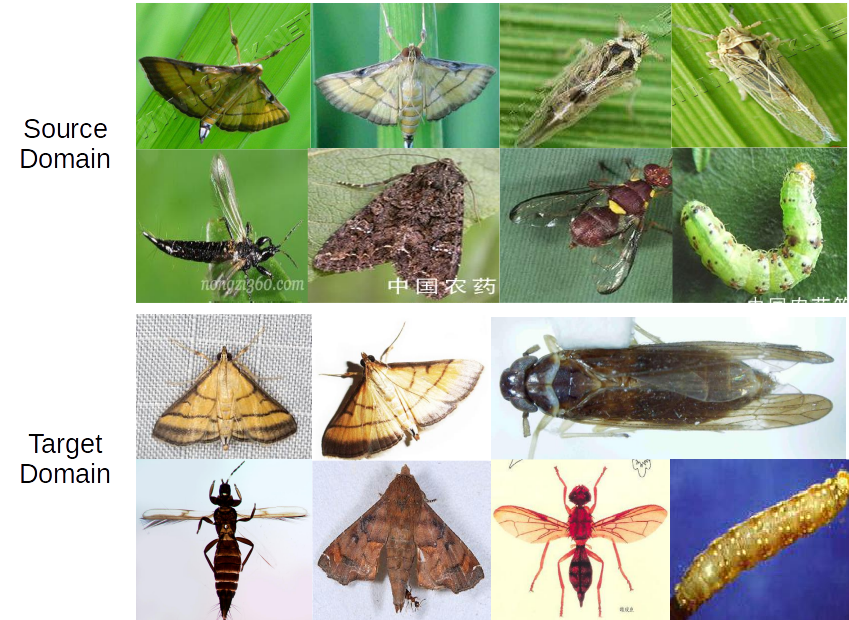}
    \caption{Examples of IP102 dataset. The images in the source domain and target domain are captured in nature and laboratories, respectively.}
    % . Meanwhile, the images in the target domain are collected in laboratories.}
    \label{fig:ip102_dataset}
    \vspace{-6mm}
\end{figure}

\begin{table}[!t]
    % \vspace{-7mm}
    \small
    \centering
    \caption{Experimental results  $(\%)$ when using SGW in various common CNNs on Insect Pest Dataset}
    \label{tab:insect_pets}
    \vspace{-3mm}
    \begin{tabular}{|c|c|c|c|}
        \hline
        \textbf{Networks} & \textbf{Methods} & \textbf{Nature} & \textbf{Laboratory} \\
        \hline
    
        \multirow{2}{*}{VGG}        & Pure CNN & 48.33\% & 47.04\% \\
        \cline{2-4}
        \multirow{2}{*}{}           &  \textbf{OTAdapt}  & \textbf{50.54\%}  & \textbf{50.35\%} \\
        \hline
        
        \multirow{2}{*}{ResNet}     & Pure CNN          & 53.05\%  & 50.96\% \\
        \cline{2-4}
        \multirow{2}{*}{}           &  \textbf{OTAdapt}  &  \textbf{55.51\%} & \textbf{53.87\%} \\
        \hline
        
        \multirow{2}{*}{DenseNet}     & Pure CNN          & 58.82\%  & 58.70\% \\
        \cline{2-4}
        \multirow{2}{*}{}           &  \textbf{OTAdapt}  &  \textbf{62.42\%} & \textbf{62.32\%} \\
        \hline
    \end{tabular}
    \vspace{-7mm}
\end{table}

\begin{table}[!b]
    \footnotesize
    \centering
    \vspace{-6mm}
    \caption{Experimental Results ($\%$) on the Office-31 Dataset (A: Amazon, W: Wecam, D: DLSR).}
    \vspace{-3mm}
    \resizebox{.5\textwidth}{!}{
    \begin{tabular}{|c|c c|c c|c c|}
        \hline
        %  Source $\to$ Target 
         &  A $\to$ W & A $\to$ D &  
                     W $\to$ A & W $\to$ D &
                     D $\to$ A & D $\to$ W  \\
        \hline
        \hline
        GFK \cite{6247911} & 58.60\% & 50.70\% & 44.10\% & 70.50\% & 45.70\% & 76.50\%  \\
        \hline
        MMDT \cite{hoffman2013efficient} &64.60\% &56.70\%&47.70\%&67.00\%&46.90\%&74.10\% \\
        \hline
        TCA \cite{TCA_method} & 72.70\% & 74.10\% & 60.90\% & $-$ & 61.70\% & $-$ \\
        \hline
        DAN \cite{DAN_method} & 78.60\% & 80.50\% &  62.80\% & $-$ &  63.60\% & $-$ \\
        \hline
        \hline
        VGG &63.64\% &71.23\% &67.21\% &65.37\% &72.54\% &68.67\%  \\
        % \hline
        % \cline{2-7}
        +\textbf{OTAdapt} &\textbf{75.32\%} &\textbf{73.83\%} &\textbf{72.37\%} &\textbf{73.69\%} &\textbf{75.48\%} &\textbf{74.57\%}  \\
        \hline
        \hline
        ResNet &61.55\% &62.44\% &74.87\% &69.23\% &71.63\% &63.80\% \\
        % \hline
        % \cline{2-7}
        +\textbf{OTAdapt} &\textbf{73.33\%} &\textbf{73.29\%} &\textbf{77.78\%} & \textbf{75.50\%} &7\textbf{8.21\%} &\textbf{73.46\%}  \\
        \hline
        \hline
        DenseNet &67.42\% &62.85\% &65.35\% &68.35\% &42.06\% &72.20\%  \\
        % \hline
        % \cline{2-7}
        +\textbf{OTAdapt} & \textbf{78.49\%} & \textbf{77.51\%} & \textbf{77.78\%} &\textbf{74.39\%} & \textbf{79.27\%} & \textbf{73.46\%} \\
        \hline
        % \hline
    \end{tabular}
    }
    \label{tab:office_exp}
    % \vspace{-6mm}
\end{table}

\textbf{IP102 Dataset:}
The IP102 dataset is a benchmark dataset for Insect Pest Recognition \cite{IP_102_dataset}.
In particular, it includes more than $75000$ images belonging to $102$ different categories collected in the Internet. In the taxonomic system of the IP102, there are 8 types of crops damaged by insect pests, specifically, Rice, Corn, Wheat, Beet, Alfalfa, Vitis, Citrus, and Mango.
Based on the property of image collection, we divide this dataset into two domains for the source and the target domains. 
The source domain is a set of images collected in nature; in particular, images were collected in the farms and outside.
Meanwhile, the target domain images are captured in laboratories.
% specifically, images are collected in the labs.
Fig. \ref{fig:ip102_dataset} illustrates the examples of the source and the target domains of the IP102 dataset.

In this experiment, the proposed method is evaluated in Insect Pest Dataset (IP102) \cite{IP_102_dataset}.
Our proposed method is evaluated with common deep network structures. In this experiment, we use image size $224\times224$, batch size and learning rate are set to $128$ and $0.0002$, respectively.
Table \ref{tab:insect_pets} shows the results of our proposed method on various deep network structures on IP102 dataset.
The experimental results in Table \ref{tab:insect_pets} show that our proposed methods help to improve the recognition performance on the target domain. Specifically, it helps to improve %\textcolor{red}
by {\textbf{$3.31\%$}, \textbf{$2.91\%$}, and $3.62\%$} on VGG, ResNet, and DenseNet, respectively.

% \textbf{Custom Dataset:}
% To further test our implementation, a custom dataset was created. This dataset is composed of $2428$ in lab bug images. This belongs to $202$ specimens of $45$ different species including but not only the Anthrax, Bombylius, and Geron families. $13$ photos per specimen were taken of which eight are from the side view and five from the top view. These photos were captured using two sources of cool light. One on the side of the specimen and another at the opposite side at about 35° from the front. Also, the lens utilized was the AF-S NIKKOR 18-55mm 1:3.5-5.6G. This allowed for shooting speeds between $1/50$s - $1/30$s, a $200$ ISO, and an f/$20$ aperture. These settings proved to be optimal to obtain a greater field of view and thus an overall in focus insect image.

% Each image is $6016\times4000$. However, due to the different specimen sizes and the lens’ focus distance the number of pixels that belongs to only the specimen varies greatly. 

\subsection{Office-31 and VisDA 2017 Experiments}

%\textcolor{red}{@RAVI: Please update the paragraph of this section again, fix grammar errors}

\textbf{Office-31 Dataset}: The Office-31 dataset is a benchmark dataset for domain adaptation \cite{saenko2010adapting}. In particular, this dataset includes 31 object categories in 3 domains i.e., Amazon , DSLR and Webcam. All the 31 categories in this dataset are the objects which are commonly seen in the  office environments. The Amazon domain contains a total of 2817 images where each class is having 90 images on average. The DSLR domain have 498 low-noise high resolution images with high resolution. Finally, for Webcam, there are a total of 795 images of low resolution with resolution of $640 \times 480$. 
% These images were captured on the online merchant's website, they also maintained uniform scale and captured against clean background. The DSLR domain have 498 low-noise high resolution images with resolution of $4288\times4288$. There are only 5 objects per category and each of them were captured from different viewpoints with an average of 3 times. Finally, for Webcam, there are a total of 795 images of low resolution with resolution of $640 \times480$. These exhibits significant noise and color as well as white balance artifacts. Few examples of the Office-31 dataset is shown in Fig. \ref{fig:office31}.

The proposed method is evaluated in Office-31 dataset \cite{saenko2010adapting}. By using the common deep neural network architectures, our proposed method is evaluated. Under this experiment, we use images of size $224\times224$, batch size is 128 and learning rate is set to 0.0001. Table \ref{tab:office_exp} shows the results of our proposed method on various deep network architectures along with baselines, i.e. Geodesic Flow Kernel (GFK) \cite{6247911}, Max-Margin Domain Transforms (MMDT) \cite{hoffman2013efficient}, TCA \cite{TCA_method}, DAN \cite{DAN_method}. This experiment demonstrates that our proposed method achieves a better recognition performance and is able to outperform the other domain adaptation techniques. 

\textbf{VisDA 2017: }
We have evaluated our approach on the VisDA dataset \cite{visda2017}. 
The source domain is a collection of synthetic images. Meanwhile, images in the target domain are real photos. We compare our results with DAN \cite{DAN_method}, DANN \cite{udab_icml}.
As shown in Table \ref{tab:visda}, our approach outperforms other baselines. Also, we conduct ablation study to illustrate the performance of our proposed components. In particular, with the adversarial loss ($\mathcal{L}_{adv}$) only, the result is $68.97\%$; Meanwhile, the Gromov-Wassertein loss only improves the result to $70.53\%$. When we use the two proposed losses together, the results have been improved up to $71.88\%$.

\begin{figure}[!t]
    \centering
    \includegraphics[width=0.40\textwidth]{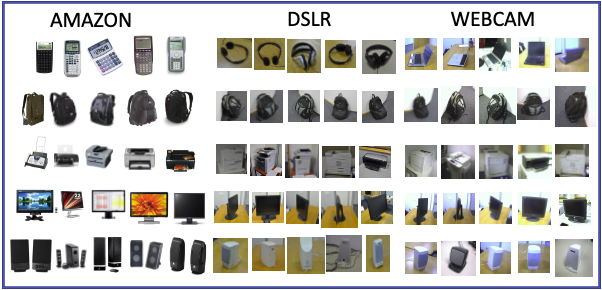}
    \vspace{-3mm}
    \caption{Examples of Office-31 Datasets} %. {\textcolor{red}{@RAVI: Please update this figure}}}
    \vspace{-6mm}
    \label{fig:office31}
\end{figure}

\begin{table}[!b]
    \centering
    \vspace{-6mm}
    \caption{Experimental Results on VisDA 2017.}
    \label{tab:visda}
    \vspace{-3mm}
    \begin{tabular}{|c|c|c|}
        \hline
        \multicolumn{2}{|c|}{Methods}         & VisDA \\ 
        \hline
        \multicolumn{2}{|c|}{Source Only}         & 52.40\%\\ 
        \hline
        \multicolumn{2}{|c|}{DAN \cite{DAN_method}} &  51.62\% \\
        \hline
        \multicolumn{2}{|c|}{DANN \cite{udab_icml}} & 57.40\% \\
        \hline
        \multirow{3}{*}{\textbf{OTAdapt}} & $\mathcal{L}_{adv}$ & 68.97\% \\
        % \hline
        \cline{2-3}
        & $\mathcal{L}_{SGW}$ & 70.53\% \\
        % \hline
        \cline{2-3}
        & $\mathcal{L}_{adv} + 
        \mathcal{L}_{SGW}$ & \textbf{71.88\%}\\
        \hline
    \end{tabular}
\end{table}

% \vspace{-5mm}
\section{Conclusions}
% \vspace{-3mm}

In this paper, we present a novel Domain Adaptation method that utilizes the optimal transport distance. 
Our proposed method is able to compare and align feature distribution across domains; meanwhile, previous methods are usually failed when the meaningful metric across domain cannot be defined. 
Through the experiment on MNIST and MNIST-M, we prove our method is able to consistently improve performance on various deep network structures and outperform other methods.
Experiments on IP102, Office-31, and VisDA have showed our method is outstanding in classification tasks.

\noindent
\textbf{Acknowledgment:}
This work is supported by NSF Small Business Innovation Research Program (SBIR), Chancellor’s Innovation Fund, and SolaRid LLC.



\bibliographystyle{IEEEtran}
\bibliography{refs}
% %
% % <OR> manually copy in the resultant .bbl file
% % set second argument of \begin to the number of references
% % (used to reserve space for the reference number labels box)


% \end{thebibliography}

% \input{rebuttal}

\end{document}